\def\year{2021}\relax
\documentclass[letterpaper]{article} % DO NOT CHANGE THIS
\usepackage{aaai21}  % DO NOT CHANGE THIS
\usepackage{times}  % DO NOT CHANGE THIS
\usepackage{helvet} % DO NOT CHANGE THIS
\usepackage{courier}  % DO NOT CHANGE THIS
\usepackage[hyphens]{url}  % DO NOT CHANGE THIS
\usepackage{graphicx} % DO NOT CHANGE THIS
\usepackage{natbib}  % DO NOT CHANGE THIS AND DO NOT ADD ANY OPTIONS TO IT
\usepackage{caption} % DO NOT CHANGE THIS AND DO NOT ADD ANY OPTIONS TO IT
\usepackage{amsmath}
\usepackage{graphicx}
\usepackage{amssymb}
\usepackage{multirow}
\usepackage{makecell}
\usepackage{mathrsfs}
\usepackage{threeparttable}
\usepackage[table,xcdraw]{xcolor}
\usepackage{booktabs}
\usepackage{color}
\usepackage{booktabs}
\usepackage{algorithm}
\usepackage{algorithmic}
\title{Deep Fusion Clustering Network}

\author{
Wenxuan Tu,$^{1,}$$\footnote{First authors with equal contribution}$ Sihang Zhou,$^{2,\ast}$ Xinwang Liu,$^{1,}$$\footnote{Corresponding author}$ Xifeng Guo,$^1$ Zhiping Cai,$^{1, \dag}$ \\En zhu,$^1$ Jieren Cheng$^{3,4}$\\
}

\affiliations{
    \textsuperscript{\rm 1}College of Computer, National University of Defense Technology, Changsha, China\\
    \textsuperscript{\rm 2}College of Intelligence Science and Technology, National University of Defense Technology, Changsha, China\\
    \textsuperscript{\rm 3}College of Computer Science and Cyberspace Security, Hainan University, Haikou, China\\
    \textsuperscript{\rm 4}Hainan Blockchain Technology Engineering Research Center, Haikou, China\\
    \{wenxuantu, guoxifeng1990, cjr22\}@163.com, sihangjoe@gmail.com, \{xinwangliu, zpcai, enzhu\}@nudt.edu.cn
}

\begin{document}

\maketitle
\begin{abstract}
Deep clustering is a fundamental yet challenging task for data analysis. Recently we witness a strong tendency of combining autoencoder and graph neural networks to exploit structure information for clustering performance enhancement. However, we observe that existing literature 1) lacks a dynamic fusion mechanism to selectively integrate and refine the information of graph structure and node attributes for consensus representation learning; 2) fails to extract information from both sides for robust target distribution (i.e., ``groundtruth" soft labels) generation. To tackle the above issues, we propose a \textbf{D}eep \textbf{F}usion \textbf{C}lustering \textbf{N}etwork (\textbf{DFCN}). Specifically, in our network, an interdependency learning-based Structure and Attribute Information Fusion (SAIF) module is proposed to explicitly merge the representations learned by an autoencoder and a graph autoencoder for consensus representation learning. Also, a 
reliable target distribution generation measure and a triplet self-supervision strategy, which facilitate cross-modality information exploitation, are designed for network training. Extensive experiments on six benchmark datasets have demonstrated that the proposed DFCN consistently outperforms the state-of-the-art deep clustering methods. \textit{Our code is publicly available at https://github.com/WxTu/DFCN.}
\end{abstract}

\section{Introduction}
Deep clustering, which aims to train a neural network for learning discriminative feature representations to divide data into several disjoint groups without intense manual guidance, is becoming an increasingly appealing direction to the machine learning researchers.
Thanks to the strong representation learning capability of deep learning methods, researches in this field have achieved promising performance in many applications including anomaly detection \cite{2020Graph}, social network analysis \cite{social}, and face recognition \cite{Facial}.
Two important factors, i.e., the optimization objective and the fashion of feature extraction, significantly determine the performance of a deep clustering method.
Specifically, in the unsupervised clustering scenario, without the guidance of labels, designing a subtle objective function and an elegant architecture to enable the network to collect more comprehensive and discriminative information for intrinsic structure revealing is extremely crucial and challenging.

\begin{figure}[!t]
\centering
\includegraphics[width=3.2in]{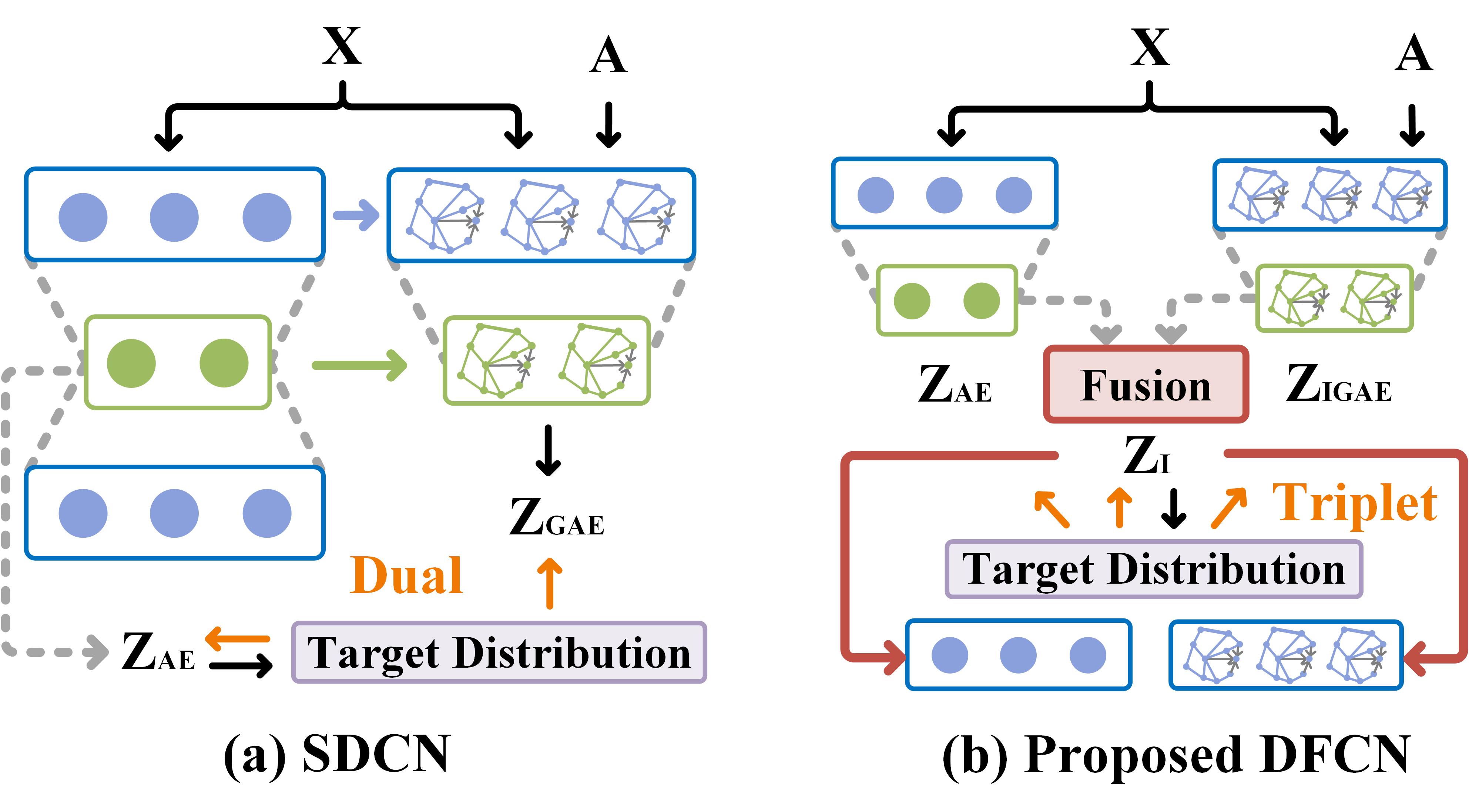}
% where an .eps filename suffix will be assumed under latex,
% and a .pdf suffix will be assumed for pdflatex; or what has been declared
% via \DeclareGraphicsExtensions.
\caption{
Network structure comparison. Different from the existing structure and attribute information fusion networks (such as SDCN), our proposed method is enhanced with an information fusion module. With this module, 1) both the decoder of AE and IGAE reconstruct the inputs with a learned consensus latent representation. 2) The target distribution is constructed with sufficient negotiation between AE and IGAE. 3) A self-supervised triplet learning strategy is designed.}
%  Our method tries to introduce modality- specific features based on the cross-modality near neighbor affinity modeling, effectively utilizing both shared and specific information for each sample.
%SDCN (a) generates the pseudo labels from one-side information, making the guidance of network training less comprehensive and accurate. While the proposed DFCN (b) designs a dynamic fusion module to adaptively integrate the information from both sides for robust pseudo label construction, and refines the latent embedding in turn through triplet learning mechanism.
\label{0}
\end{figure}

According to the network optimization objective, existing deep clustering methods can be roughly grouped into five categories, i.e., subspace clustering-based methods~\cite{LDPDSC, DSCN, CSC}, generative adversarial network-based methods~\cite{2019ClusterGAN, BSL}, spectral clustering-based methods~\cite{DSCUAN, SpectralNet}, Gaussian mixture model-based methods~\cite{GMMGE, UCQIP}, and self-optimizing-based methods~\cite{2015Unsupervised, 2017Improved}. Our method falls into the last category.
In the early state, the above deep clustering methods mainly concentrate on exploiting the attribute information in the original feature space of data and have achieved good performance in many circumstances. To further improve the clustering accuracy, recent literature shows a strong tendency in extracting geometrical structure information and then integrates it with attribute information for representation learning.
Specifically, Yang et al. design a novel stochastic extension of graph embedding to add local data structures into probabilistic deep Gaussian mixture model (GMM) for clustering \cite{GMMGE}. Distribution preserving subspace clustering (DPSC) first estimates the density distribution of the original data space and the latent embedding space with kernel density estimation. Then it preserves the intrinsic cluster structure within data by minimizing the distribution inconsistency between the two spaces \cite{LDPDSC}. More recently, graph convolutional networks (GCNs), which aggregate the neighborhood information for better sample representation learning, have attracted the attention of many researchers. The work in deep attentional embedded graph clustering (DAEGC) exploits both graph structure and node attributes with a graph attention encoder. It reconstructs the adjacency matrix by a self-optimizing embedding method \cite{DAEGC}. Following the setting of DAEGC, adversarially regularized graph autoencoder (ARGA) further develops an adversarial regularizer to guide the learning of latent representations~\cite{Pan2019Learning}.
After that, structural deep clustering network (SDCN) \cite{Bo2020Structural} integrates an autoencoder and a graph convolutional network into a unified framework by designing an information passing delivery operator and a dual self-supervised learning mechanism.

Although the former efforts have achieved preferable performance enhancement by leveraging both kinds of information, we find that 1) the existing methods lack an cross-modality dynamic information fusion and processing mechanism. Information from two sources is simply aligned or concatenated, leading to insufficient information interaction and merging; 2) the generation of the target distribution in existing literature has seldom used information from both sources, making the guidance of network training less comprehensive and accurate. As a consequence, the negotiation between two information sources is obstructed, resulting in unsatisfying clustering performance.

%i) these methods only utilize one-side information in constructing pseudo labels, which is crucial to guide the unsupervised network learning. This not only ignores the other regularized information on learning consensus and discriminative representations, but also prevents the complementary features from being utilized; ii) none of them sufficiently considers a cross-modality dynamic fusion mechanism to adaptively integrate the information of node attributes and graph structure. This overlooks the possibility that two modality information may need to negotiate with each other for revealing the intrinsic structure relations. Both limitations could adversely affect the robustness of deep embedding, resulting in unsatisfying clustering performance.

% Inspired by the recent developments in attributed graph clustering filed

To tackle the above issues, we propose a deep fusion clustering network (DFCN). The main idea of our solution is to design a dynamic information fusion module to finely process the attribute and structure information extracted from autoencoder (AE) and graph autoencoder (GAE) for a more comprehensive and accurate representation construction. Specifically, a structure and attribute information fusion (SAIF) module is carefully designed for elaborating both-source information processing. Firstly, we integrate two kinds of sample embeddings in both the perspective of local and global level for consensus representation learning. After that, by estimating the similarity between sample points and pre-calculated cluster centers in the latent embedding space with Students' \textit{t}-distribution, we acquire more precise target distribution. Finally, we design a triplet self-supervision mechanism which uses the target distribution to provide more dependable guidance for AE, GAE, and the information fusion part simultaneously. Moreover, we develop an improved graph autoencoder (IGAE) with a symmetric structure and reconstruct the adjacency matrix with both the latent representations and the feature representations reconstructed by the graph decoder. The key contributions of this paper are listed as follows:

%reconstructing the adjacency matrix to update the latent embedding, we incorporate the graph reconstruction that acts as graph-related regularization term into the optimization objective, in order to serve better for the cross-modality dynamic fusion process and the robust embedding learning. The key contributions of this paper are three-fold:

%To achieve this goal, the proposed Structure and Attribute Information Fusion (SAIF) jointly encodes shared-specific information of node attributes and graph structure to generate robust pseudo labels. Specifically, we design a triplet self-supervision strategy where pseudo labels and soft clustering assignments are uniformly aligned, thereby iteratively facilitating cross-modality information exploitation in turn. Thereafter, a cross-modality dynamic fusion mechanism is presented to build a bridge between autoencoder (AE) and graph autoencoder (GAE) in a group-wise manner. By this means, more potential connected information among two-modality samples will be explored to negotiate with each other for learning the consensus and discriminative representations. This contributes to achieve mutual benefit for AE and GAE.

\begin{itemize}
%\item A novel deep clustering method named DFCN is proposed. To the best of our knowledge, we propose the first clustering method that can adaptively leverage both attributed features and graph structure from fusion-based autoencoders.

\item We propose a deep fusion clustering network (DFCN). In this network, a structure and attribute information fusion (SAIF) module is designed for better information interaction between AE and GAE. With this module, 1) since both the decoders of AE and GAE reconstruct the inputs using a consensus latent representation, the generalization capacity of the latent embeddings is boosted. 2) The reliability of the generated target distribution is enhanced by integrating the complementary information between AE and GAE. 3) The self-supervised triplet learning mechanism integrates the learning of AE, GAE and the fusion part in a unified and robust system, thus further improves the clustering performance.

%Besides the proposal of the SAIF module, we also develop a symmetric graph autoencoder to further improve the generalization capability of the proposed method.
\item We develop a symmetric graph autoencoder, i.e., improved graph autoencoder (IGAE), to further improve the generalization capability of the proposed method.

%Then, a triplet learning mechanism is proposed to provide precise network training guidance. Moreover, we further develop a novel symmetric graph autoencoder to improve the generalization capability of the proposed method.

%We design a SAIF module for better representation integration and more accurate pseudo label construction, which involves a cross-modality dynamic fusion mechanism and a triplet learning strategy to finely integrate the information from both sources and provide precise network training guidance, respectively.

%In the method, a structure and attribute information fusion (SAIF) module is designed for better representation integration and more accurate pseudo label construction. Then, a triplet learning mechanism is proposed to provide precise network training guidance. Moreover, we further develop a novel symmetric graph autoencoder to improve the generalization capability of the proposed method.

\item Extensive experiment results on six public benchmark datasets have demonstrated that our method is highly competitive and consistently outperforms the state-of-the-art ones with a preferable margin.
\end{itemize}
%We present a Structure and Attribute Information Fusion (SAIF) module to reveal the intrinsic structure relations and generate the robust pseudo labels for clustering performance improvement by a cross-modality dynamic fusion mechanism and a triplet self-supervision strategy.
%which can mutually exploit node attributes and graph structure to learn consensus and discriminative representations for better clustering.
%\item

\section{Related Work}
%In this section, we briefly discuss two lines of related work, attributed graph clustering and self-supervised clustering, which are most relevant to our work.

\subsection{Attributed Graph Clustering}
Benefiting from the strong representation power of graph convolutional networks (GCNs) \cite{2016Semi}, GCN-based clustering methods that jointly learn graph structure and node attributes have been widely studied in recent years~\cite{OGA, MAFCN, MSL}. 
Specifically, graph autoencoder (GAE) and variational graph autoencoder (VGAE) are proposed to integrate graph structure into node attributes via iteratively aggregating neighborhood representations around each central node~\cite{2016Variational}. After that, ARGA~\cite{Pan2019Learning}, AGAE~\cite{2019Adversarial}, DAEGC~\cite{DAEGC}, and MinCutPool~\cite{2020SCGNN} improve the performance of the early-stage methods with adversarial training, attention, and graph pooling mechanisms, respectively. Although the performance of the corresponding methods has been improved considerably, the over-smoothing phenomenon of the GCNs still limits the accuracy of these methods.
More recently, SDCN~\cite{Bo2020Structural} is proposed to integrate autoencoder and GCN module for better representation learning. Through careful theoretical and experimental analysis, authors find that in their proposed network, autoencoder can help provide complementary attribute information and help relieve the over-smoothing phenomenon of GCN module, while GCN module provides high-order structure information to autoencoder. Although SDCN proves that combining autoencoder and GCN module can boost the clustering performance of both components, in this work, the GCN module acts only as a regularizer of the autoencoder. Thus, the learned features of the GCN module are insufficiently utilized for guiding the self-optimizing network training and the representation learning of the framework lacks the negotiation between the two sub-networks. Differently, in our proposed method, an information fusion module (i.e., SAIF module) is proposed to integrate and refine the features learned by the AE and IGAE. As a consequence, the complementary information from two sub-networks is finely merged to reach a consensus, and more discriminative representations are learned.

%efforts have been made on designing a cross-modality clustering network that manages to integrate the characteristic of the data itself into graph information via a delivery operation for the first time \cite{Bo2020Structural}. Though carefully exploiting localized neighbor relations underlying data, methods stated above only capture shallow relationships between structure and attribute information. Thus, these methods lack a dynamic fusion mechanism to adaptively integrate two-modality representations, resulting in under-utilizing the interplay between graph structure and node attributes.
%Following the same strategy, ARGA \cite{Pan2019Learning} and AGAE \cite{2019Adversarial} encode both graph structure and node attributes into a compact representation, and model the interaction between them to guide the latent embedding learning with a adversarial module. Similarly, DAEGC \cite{DAEGC} designs a two-layer attention-based GCN to exploit the importance of neighbor node attributes, then recovers adjacency matrix with a simple inner product decoder. MinCutPool \cite{2020SCGNN} applies spectral clustering on both the underlying graph structure and node attributes to achieve a consistent partition. 

\subsection{Target Distribution Generation}
Since reliable guidance is missing in clustering network training, many deep clustering methods seek to generate the target distribution (i.e., ``groundtruth" soft labels) for discriminative representation learning in a self-optimizing manner~\cite{2019Semi, AIMC, DAMCN}.
The early method (DEC) in this category first trains an encoder, and then with the pre-trained network, it further defines a target distribution based on the Student's \textit{t}-distribution and fine-tunes the network with stronger guidance~\cite{2015Unsupervised}. To increase the accuracy of the target distribution, IDEC jointly optimizes the cluster assignment and learns features that are suitable for clustering with local structure preservation~\cite{2017Improved}. 
%Subsequently, DAEGC \cite{DAEGC} and AGAE \cite{2019Adversarial} propose a structure-based autoencoder to learn graph representations that generate and update pseudo labels with attention guided graph convolutions and adversarial regularizer, respectively. 
After that, to better train the autoencoder and GCN module integrated network, SDCN designs a dual self-supervised learning mechanism which conducts target distribution refinement and sub-network training in a unified system~\cite{Bo2020Structural}.
Despite their success, existing methods generate the target distribution with only the information of autoencoder or GCN module. None of them considers combining the information from both sides and then comes up with a more robust guidance, thus the generated target distribution could be less comprehensive. 
In contrast, in our method, as the information fusion module allows the information from the two sub-networks to adequately interact with each other, the resultant target distribution has the potential to be more reliable and robust than that of the single-source counterparts.

\section{The Proposed Method}
%In this section, we first formalize the deep clustering task on attributed graphs. Then we introduce each component of the proposed the framework. Fig. \ref{0} shows an overview of DFCN that mainly consists of four components: AE, IGAE, a dynamic fusion module, and the optimization strategy. Our AE adopts a common symmetric structure~\cite{Bo2020Structural}, of which details would be illustrated in supplementary materials. We will elaborate the design of IGAE and SAIF module as follows. Finally we analyze the optimization objective and training procedure in detail. 

Our proposed method mainly consists of four parts, i.e., an autoencoder, an improved graph autoencoder, a fusion module, and the optimization targets (please check Fig. \ref{0} for the diagram of our network structure). The encoder part of both AE and IGAE are similar with that of the existing literature. 
In the following sections, we will first introduce the basic notations and then introduce the decoder of both sub-networks, the fusion module, and the optimization targets in detail.

\subsection{Notations}
Given an undirected graph $\mathcal{G}=\{\mathcal{V, E}\}$ with $K$ cluster centers, $\mathcal{V} = \{v_{1}, v_{2}, \dots, v_{N}\}$ and $E$ are the node set and the edge set, respectively,  where $N$ is the number of samples. The graph is characterized by its attribute matrix $\mathbf{X} \in \mathbb{R}^{N \times d}$ and original adjacency matrix $\mathbf{A}=(a_{ij})_{N \times N} \in \mathbb{R}^{N \times N}$. Here, $d$ is the attribute dimension and $a_{ij}=1$ if $(v_{i}, v_{j}) \in \mathcal{E}$, otherwise $a_{ij} = 0$.
%The structure of graph \textit{G} can be described as an adjacency matrix \textit{A} = \textit{$\{a_{ij}\}$} $\in$ $\mathbb{R}$$^{N \times N}$, where \textit{a$_{ij}$} = 1 if $\left(\textit{v$_{i}$, v$_{j}$}\right)$ $\in$ \textit{E}, otherwise \textit{a$_{ij}$} = 0. 
The corresponding degree matrix is $\mathbf{D}= diag(d_{1}, d_{2}, ..., d_{N}) \in \mathbb{R}^{N \times N}$  and $d_{i} = \sum_{v_{j} \in V}a_{ij}$. With $\mathbf{D}$, the original adjacency matrix is further normalized as $\widetilde{\mathbf{A}} \in \mathbb{R}^{N \times N}$ through calculating $\mathbf{D}^{-\frac{1}{2}}(\mathbf{A} + \mathbf{I}) \mathbf{D}^{-\frac{1}{2}}$, where $\mathbf{I} \in \mathbb{R}^{N \times N}$  indicates that each node in $\mathcal{V}$ is linked with a self-loop structure. All notations are summarized in Table \ref{I}.

\begin{table}[!t]
\centering

\footnotesize
%\label{tab:my-table}
\begin{tabular}{l|l}\hline
\hline
\multicolumn{1}{c|}{Notations} & \multicolumn{1}{c}{Meaning}   \\\hline
$\mathbf{X} \in \mathbb{R}^{N \times d}$    & Attribute matrix   \\
$\mathbf{A} \in \mathbb{R}^{N \times N}$    & Original adjacency matrix   \\
$\mathbf{I} \in \mathbb{R}^{N \times N}$    & Identity matrix   \\
$\widetilde{\mathbf{A}} \in \mathbb{R}^{N \times N}$    & Normalized adjacency matrix   \\
$\mathbf{D} \in \mathbb{R}^{N \times N}$ & Degree matrix \\

$\widehat{\mathbf{Z}} \in \mathbb{R}^{N \times d}$ & Reconstructed weighted attribute matrix\\
$\widehat{\mathbf{A}} \in \mathbb{R}^{N \times N}$ & Reconstructed adjacency matrix \\
$\mathbf{Z}_{AE} \in \mathbb{R}^{N \times d^{'}}$ & Latent embedding of AE\\
$\mathbf{Z}_{IGAE} \in \mathbb{R}^{N \times d^{'}}$ & Latent embedding of IGAE \\
$\mathbf{Z}_I \in \mathbb{R}^{N \times d^{'}}$ & Initial fused embedding \\
$\mathbf{Z}_L \in \mathbb{R}^{N \times d^{'}}$ & Local structure enhanced $\mathbf{Z}_I$ \\
$\mathbf{S} \in \mathbb{R}^{N \times N}$ & Normalized self-correlation matrix  \\

$\mathbf{Z}_G \in \mathbb{R}^{N \times d^{'}}$ & Global structure enhanced $\mathbf{Z}_L$  \\
$\widetilde{\mathbf{Z}}\in \mathbb{R}^{N \times d^{'}}$ & Clustering embedding  \\
$\mathbf{Q}\in \mathbb{R}^{N \times K}$ & Soft assignment distribution  \\
$\mathbf{P}\in \mathbb{R}^{N \times K}$ & Target distribution  \\

\hline\hline
\end{tabular}
\caption{Basic notations for the proposed DFCN}
\label{I}
\end{table}

%\textit{D} = \textit{diag$\left(d_{1}, d_{2}, ..., d_{N}\right)$}, where \textit{d$_{i}$} = $\sum_{v_{j} \in V}$\textit{a$_{ij}$} denotes the degree of node \textit{v$_{i}$}, and the graph normalized adjacency matrix is presented as \textit{D$^{-\frac{1}{2}}$$\widetilde{A}$D$^{-\frac{1}{2}}$}, where \textit{$\widetilde{A}$ = A + I} indicates that each node contains self-loop information.  Our goal is to generate the robust embedding matrix, on which we partition all nodes into \textit{k} disjoint groups $\{\textit{G$_{1}$, G$_{2}$, ..., G$_{k}$}\}$.

%\subsection{Overall Framework}
%ig. \ref{0} shows an overview of our framework that mainly consists of four components: AE, IGAE, a dynamic fusion module, and the optimization strategy. Our AE adopts a common symmetric structure~\cite{Bo2020Structural}, of which details would be illustrated in supplementary materials. We will elaborate the design of IGAE, SAIF module, and optimization strategy as follows.
\subsection{Fusion-based Autoencoders}
\noindent{\textbf{Input of the Decoder}}.
%\subsection{Fusion-based Autoencoders}
Most of the existing autoencoders, either classic autoencoder or graph autoencoder, reconstruct the inputs with only its own latent representations. However, in our proposed method, with the compressed representations of AE and GAE, we first integrate the information from both sources for a consensus latent representation. Then, with this embedding as an input, both the decoders of AE and GAE reconstruct the inputs of two sub-networks. This is very different from the existing methods that our proposed method fuses heterogeneous structure and attribute information with a carefully designed fusion module and then reconstructs the inputs of both sub-networks with the consensus latent representation. Detailed information about the fusion module will be introduced in the Structure and Attribute Information Fusion section.

\noindent{\textbf{Improved Graph Autoencoder}}.
%\subsection{Improved Graph Autoencoder}
In the existing literature, the classic autoencoders are usually symmetric, while graph convolutional networks are usually asymmetric~\cite{2016Variational, DAEGC, 2019Adversarial}. They require only the latent representation to reconstruct the adjacency information and overlook that the structure-based attribute information can also be exploited for improving the generalization capability of the corresponding network. To better make use of both the adjacency information and the attribute information, we design a symmetric improved graph autoencoder (IGAE). This network requires to reconstruct both the weighted attribute matrix and the adjacency matrix simultaneously. In the proposed IGAE, a layer in the encoder and decoder is formulated as:

%Attributed graph clustering methods usually update the latent embedding by reconstructing the adjacency matrix, which only records structure information and ignores other regularized information \cite{2016Variational, DAEGC, 2019Adversarial}. To this end, we propose IGAE to enrich the diversity of network representations.
%Given node features \textit{X} and graph structure \textit{A}, graph representations in encoder and decoder can be formulated as:

\begin{equation} \label{eq:1}
\mathbf{Z}^{(l)} = \sigma(\widetilde{\mathbf{A}}\mathbf{Z}^{(l-1)}\mathbf{W}^{(l)}),
\end{equation}

\begin{equation} \label{eq:2}
\mathbf{\widehat{Z}}^{(h)} = \sigma(\widetilde{\mathbf{A}}\widehat{\mathbf{Z}}^{(h-1)}\widehat{\mathbf{W}}^{(h)}),
\end{equation}
where $\mathbf{W}^{(l)}$ and $\mathbf{\widehat{W}}^{(h)}$ denote the learnable parameters of the \textit{l}-th encoder layer and \textit{h}-th decoder layer. $\sigma$ is a non-linear activation function, such as ReLU or Tanh.
%The input of the first layer \textit{Z$^{(0)}$} and \textit{$\widetilde{Z}$$^{(0)}$} refer to the original feature matrix \textit{X} and the latent embedding learned from graph encoder.
%To clearly reveal the intrinsic structure relations of graph, we adopt two simple inner product layers that serve as constraint at the end of graph encoder and decoder, respectively. Thus, we obtain the generated adjacency matrix as follows :
%\begin{equation} \label{eq:3}
%\widehat{A} = sigmoid(ZZ^{T}) + sigmoid(\widetilde{Z}\widetilde{Z}^{T}),
%\end{equation}
%where \textit{Z} and \textit{$\widetilde{Z}$} refer to the graph embedding matrix and reconstructive graph matrix, respectively.
To minimize both the reconstruction loss functions over the weighted attribute matrix and the adjacency matrix, our IGAE is designed to minimize a hybrid loss function:
\begin{equation} \label{eq:6}
L_{IGAE} = L_{w} + \gamma L_{a}.
\end{equation}
In Eq.\eqref{eq:6}, $\gamma$ is a pre-defined hyper-parameter that balances the weight of the two reconstruction loss functions. Specially, $L_w$ and $L_a$ are defined as follows:

\begin{equation} \label{eq:5}
L_w = \frac{1}{2N}\Arrowvert\widetilde{\mathbf{A}}\mathbf{X} - \widehat{\mathbf{Z}}\Arrowvert_F^2,
\end{equation}

\begin{equation} \label{eq:4}
L_a= \frac{1}{2N}\Arrowvert\widetilde{\mathbf{A}} - \widehat{\mathbf{A}}\Arrowvert_F^2.
\end{equation}
In Eq.\eqref{eq:5}, $\widehat{\mathbf{Z}} \in \mathbb{R}^{N \times d}$ is the reconstructed weighted attribute matrix. In Eq.\eqref{eq:4}, $\widehat{\mathbf{A}} \in \mathbb{R}^{N \times N}$ is the reconstructed adjacency matrix generated by an inner product operation with multi-level representations of the network. By minimizing both Eq.\eqref{eq:5} and Eq.\eqref{eq:4}, the proposed IGAE is termed to minimize the reconstruction loss over the weighted attribute matrix and the adjacency matrix at the same time. Experimental results in the following parts validate the effectiveness of this setting.

%Following the previous work ~\cite{2016Variational}, we minimize the reconstruction loss \textit{L$_{s}$} between the normalized adjacency matrix and the one generated by the measure of inner product decoder:
%\begin{equation} \label{eq:4}
%L_s= \frac{1}{2N}\Arrowvert D^{-\frac{1}{2}}\widetilde{A}D^{-\frac{1}{2}}- \widehat{A}\Arrowvert_F^2.
%\end{equation}

%Apart from optimizing the structure reconstruction error, we consider that incorporate the reconstruction of graph itself as regularized term into the optimization objective, as formulated by Eq.(\ref{eq:5}). Finally, both graph-based and structure-based loss function are jointly optimized during the training process according to Eq.(\ref{eq:6}). By this means, IGAE exhibits two advantages compared with the proposal that  reconstructs sparse adjacency matrix only, 1) boosting the quality of network representations via introducing extra regularized term into the optimization process; 2) The latent embedding could inherit more properties from the feature space of the original graph, preserving representative meaningful features that generate better clustering decisions. 

%\begin{equation} \label{eq:5}
%L_g = \frac{1}{2N}\Arrowvert D^{-\frac{1}{2}}\widetilde{A}D^{-\frac{1}{2}}X- \widetilde{Z}\Arrowvert_F^2,
%\end{equation}

%\begin{equation} \label{eq:6}
%L_{IGAE} = L_g + \gamma L_s,
%\end{equation}
%where $\gamma$ = 0.1 is a hyper-parameter that controls the disturbance of structure space to the graph embedding space.
\begin{figure}[!t]
\centering
\includegraphics[width=3.3in]{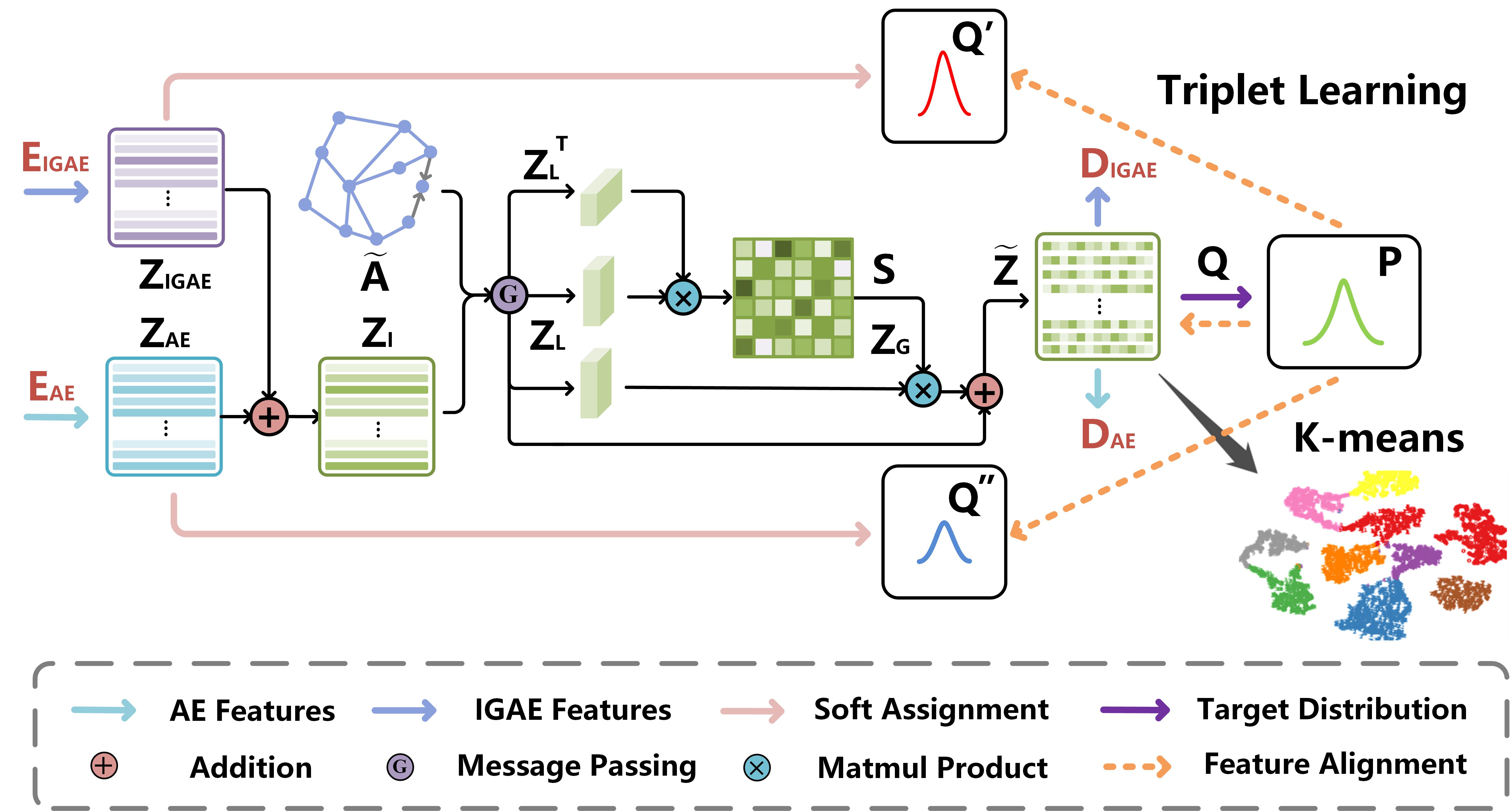}
% where an .eps filename suffix will be assumed under latex,
% and a .pdf suffix will be assumed for pdflatex; or what has been declared
% via \DeclareGraphicsExtensions.
\caption{ Illustration of the Structure and Attribute Information Fusion (SAIF) module.}
\label{1}
\end{figure}

\subsection{Structure and Attribute Information Fusion}
%To sufficiently explore the node attributes, graph structure or both in AE and GAE, a direct solution is to design a complementary feature fusion mechanism that mutually exploits modality information. Our proposal follows the same strategy of SDCN but differs from it that SAIF module adaptively integrates the node attributes and graph structure in a dynamic interactive manner, and facilitates the optimization of shared-specific information via a triplet self-supervised strategy.
%SDCN designs a delivery operator to integrate node attributes learned by DNN into graph information learned by GCN for the first time \cite{Bo2020Structural}. 
To sufficiently explore the graph structure and node attributes information extracted by the AE and IGAE, we propose a structure and attribute information fusion (SAIF) module. This module consists of two parts, i.e., a cross-modality dynamic fusion mechanism and a triplet self-supervised strategy. The overall structure of SAIF is illustrated in Fig. \ref{1}.

\subsubsection{Cross-modality Dynamic Fusion Mechanism.}
The information integration within our fusion module includes four steps.
First, we combine the latent embedding of AE ($\mathbf{Z}_{AE} \in \mathbb{R}^{N \times d^{'}}$) and IGAE ($\mathbf{Z}_{IGAE} \in \mathbb{R}^{N \times d^{'}}$) with a linear combination operation:

\begin{equation} \label{eq:9}
\mathbf{Z}_I = \alpha \mathbf{Z}_{AE}+(1-\alpha)\mathbf{Z}_{IGAE},
\end{equation}
where $d^{'}$ is the latent embedding dimension, and $\alpha$ is a learnable coefficient which selectively determines the importance of two information sources according to the property of the corresponding dataset. In our paper, $\alpha$ is initialized as $0.5$ and then tuned automatically with a gradient decent method.

Then, we process the combined information with a graph convolution-like operation (i.e., message passing operation). 
%In the convolution, the adjacent matrix is the self-looped adjacent matrix. 
With this operation, we enhance the initial fused embedding $\mathbf{Z}_I \in \mathbb{R}^{N \times d^{'}}$ by considering the local structure within data: 
\begin{equation} \label{eq:10}
\mathbf{Z}_L = \widetilde{\mathbf{A}}\mathbf{Z}_I.
\end{equation}   
In Eq.\eqref{eq:10}, $\mathbf{Z}_L \in \mathbb{R}^{N \times d^{'}}$ denotes the local structure enhanced $\mathbf{Z}_I$.

After that, we further introduce a self-correlated learning mechanism to exploit the non-local relationship in the preliminary information fusion space among samples. Specifically, we first calculate the normalized self-correlation matrix $\mathbf{S} \in \mathbb{R}^{N \times N}$ through Eq.\eqref{eq:11}: 
\begin{equation} \label{eq:11}
\mathbf{S}_{ij} = \frac{e^{({\mathbf{Z}_L\mathbf{Z}_L^{\mathbf{T}}})_{ij}}}{\sum_{k=1}^N e^{({\mathbf{Z}_L\mathbf{Z}_L^{\mathbf{T}}})_{ik}}}.
\end{equation}
With $\mathbf{S}$ as coefficients, we recombine $\mathbf{Z}_L$ by considering the global correlation among samples: $\mathbf{Z}_G = \mathbf{S}\mathbf{Z}_L$.

Finally, we adopt a skip connection to encourage information to pass smoothly within the fusion mechanism:
\begin{equation} \label{eq:12}
\widetilde{\mathbf{Z}} = \beta \mathbf{Z}_G + \mathbf{Z}_L,
\end{equation} 
where $\beta$ is a scale parameter. Following the setting in \cite{DANet}, we initialize it as 0 and learn its weight while training the network. 
Technically, our cross-modality dynamic fusion mechanism considers the sample correlation in both the perspective of the local and global level. Thus, it has potential benefit on finely fusing and refining the information from both AE and IGAE for learning consensus latent representations.

%As depicted in Fig. \ref{1}, we further adopt dot-product similarity to measure the affinities between samples, and apply a softmax layer to calculate the similarity score matrix $\mathbf{S} \in \mathbb{R}^{N \times N}$. Now, we have defined a similarity graph that more similar feature representations of two nodes contributes to greater correlation between them, whether they are neighbors or not. Meanwhile, we employ a matrix multiplication between $\mathbf{\widetilde{F}}$ and $\mathbf{S}$, then perform a skip connection operation with original $\mathbf{\widetilde{F}}$ to obtain the final deep hybrid embedding $\mathbf{\widehat{F}} \in \mathbb{R}^{N \times d}$, which will be fed into both decoders of AE and IGAE eventually. By this means, the rich group-wise relation information among different node pairs are involved, which will guide the refinement of network representations from both AE and IGAE. Thus, sufficient node attributes and graph structure would be adaptively explored to assist in generating more robust and accurate pseudo labels. The above process can be expressed as:

%where $\otimes$ and $\oplus$ refer to element-wise multiplication and element-wise addition, respectively. $\beta$ is initialized as 0 and gradually learns to assign more weight \cite{DANet}.

\subsubsection{Triplet Self-supervised Strategy.}
To generate more reliable guidance for clustering network training, we first adopt the more robust clustering embedding $\widetilde{\mathbf{Z}} \in \mathbb{R}^{N \times d^{'}}$ which has integrated the information from both AE and IGAE for target distribution generation. As shown in Eq.\eqref{eq:13} and Eq.\eqref{eq:14}, the generation process includes two steps:
\begin{equation} \label{eq:13}
q_{ij} = \frac{(1+\Arrowvert \tilde{z}_{i} - u_{j} \Arrowvert^2/v)^{-\frac{v+1}{2}}}{\sum_{j^{'}}(1+\Arrowvert \tilde{z}_{i} - u_{j^{'}} \Arrowvert^2/v)^{-\frac{v+1}{2}}},
\end{equation}

\begin{equation} \label{eq:14}
p_{ij} = \frac{q^{2}_{ij}/\sum_{i}q_{ij}}{\sum_{j^{'}}(q^{2}_{ij^{'}}/\sum_{i}q_{ij^{'}})}.
\end{equation}
In the first step (corresponding to Eq.\eqref{eq:13}), we calculate the similarity between the $i$-th sample ($\tilde{z}_{i}$) and the $j$-th pre-calculated clustering center ($u_j$) in the fused embedding space using Student's \textit{t}-distribution as kernel. In Eq.\eqref{eq:13}, $v$ is the degree of freedom for Student's \textit{t}-distribution and $q_{ij}$ indicates the probability of assigning the \textit{i}-th node to the \textit{j}-th center (i.e., a soft assignment). The soft assignment matrix $\mathbf{Q} \in \mathbb{R}^{N \times K}$ reflects the distribution of all samples. In the second step, to increase the confidence of cluster assignment, we introduce Eq.\eqref{eq:14} to drive all samples to get closer to cluster centers. Specifically, $0 \leq p_{ij} \leq 1$ is an element of the generated target distribution $\mathbf{P} \in \mathbb{R}^{N \times K}$, which indicates the probability of the $i$-th sample belongs to the $j$-th cluster center.

%Deep clustering task generally faces a common challenge that how to provide guidance information to define clustering-oriented optimization objective. Different from previous single or dual self-supervised measures, we develop a triplet one as a solution. Following Student's \textit{t}-distribution method~\cite{2008Visualizing}, we first obtain several soft assignment distribution $\mathbf{Q}$, $\mathbf{Q^{'}}$ and $\mathbf{Q^{''}}$ calculated by $\mathbf{\widehat{Z}}$, $\mathbf{Z}_{AE}^{(m)}$, and $\mathbf{Z}_{IGAE}^{(n)}$, respectively:
%where \textit{v} denotes the number of degrees of freedom for Student's \textit{t}-distribution. \textit{q$_{ij}$} refers to the probability of assigning \textit{i}-th node to \textit{j}-th centroid. \textit{u} refers to initial clustering centroids. Subsquently, we calculate the target distribution (i.e., pseudo labels) $\mathbf{P}$ derived from $\mathbf{Q}$ to make $\mathbf{\widehat{Z}}$ closer to cluster centroids:
%where $\sum_{i}$\textit{q$_{ij}$} denotes soft cluster frequencies, and the soft assignment $\mathbf{Q}$ is squared and normalized to emphasize confident assignments.
With the iteratively generated target distribution, we then calculate the soft assignment distribution of AE and IGAE by using Eq.\eqref{eq:13} over the latent embeddings of two sub-networks, respectively. We denote the soft assignment distribution of IGAE and AE as $\mathbf{Q}^{'}$ and $\mathbf{Q}^{''}$. 
%Both $\mathbf{q}^{'}_{ij}$ and $\mathbf{q}^{''}_{ij}$ refer to the similar concept of $\mathbf{q}_{ij}$.

To train the network in a unified framework and improve the representative capability of each component, we design a triplet clustering loss by adapting the KL-divergence in the following form:
\begin{equation} \label{eq:15}
 L_{KL}= \sum_{i}\sum_{j}p_{ij}log\frac{p_{ij}}{(q_{ij}+q^{'}_{ij}+q^{''}_{ij})/3}.
\end{equation}
In this formulation, the summation of soft assignment distribution of AE, IGAE, and the fused representations are aligned with the robust target distribution simultaneously. Since the target distribution is generated without human guidance, we name the loss function triplet clustering loss and the corresponding training mechanism as triplet self-supervised strategy.

%Our self-supervised strategy has two advantages. On the one hand, the shared soft assignment distribution $\mathbf{Q}$, which might be regarded as a optimal choice to generate the target distribution $\mathbf{P}$, is more complete and robust compared with specific ones $\mathbf{Q^{'}}$ or $\mathbf{Q^{''}}$. On the other hand, the robust generated pseudo labels that give a positive feedback for shared and specific embedding space learning would provide more dependable guidance for the optimization of AE, GAE, and information fusion simultaneously. Hence, both soft assignment distribution and target distribution could iteratively achieve mutual gains for better clustering.

\begin{algorithm}[!htbp]
{\caption{Deep Fusion Clustering Network}\label{Algorithm-proposed}
\small
\begin{algorithmic}[1]
\REQUIRE Attribute matrix $\mathbf{X}$; Adjacency matrix $\mathbf{A}$; Target distribution update interval \textit{T}; Iteration number \textit{I}; Cluster number \textit{K}; Hyper-parameters $\gamma$, $\lambda$.
\ENSURE Clustering results $\mathbf{O}$.

\STATE Initialize the parameters of AE, IGAE, and the fusion part to obtain $\mathbf{Z}_{AE}$, $\mathbf{Z}_{IGAE}$, and $\mathbf{\widetilde{Z}}$;
\STATE Initialize the clustering centers \textit{u} with K-means based on $\mathbf{\widetilde{Z}}$;
\FOR {$\textit{i} = 1$ to $\textit{I}$}
\STATE Update $\mathbf{Z}_{I}$ and $\mathbf{Z}_{L}$ by Eq.(6) and Eq.(7);
\STATE Update the normalized self-correlation matrix $\mathbf{S}$ and the deep clustering embedding $\mathbf{\widetilde{Z}}$ by Eq.(8) and Eq.(9), respectively;
\STATE Calculate soft assignment distributions $\mathbf{Q}$, $\mathbf{Q^{'}}$, and $\mathbf{Q^{''}}$ based on  $\mathbf{\widetilde{Z}}$, $\mathbf{Z}_{IGAE}$, and $\mathbf{Z}_{AE}$ by Eq.(10);
\IF {$ \textit{i} \% \textit{T} == 0$}
\STATE Calculate the target distribution $\mathbf{P}$ derived from $\mathbf{Q}$ by Eq.(11);
\ENDIF
\STATE Utilize $\mathbf{P}$ to refine $\mathbf{Q}$, $\mathbf{Q^{'}}$, and $\mathbf{Q^{''}}$ in turn by Eq.(12);
\STATE Calculate \textit{L$_{AE}$}, \textit{L$_{IGAE}$}, and \textit{L$_{KL}$}, respectively.
\STATE Update the whole network by minimizing Eq.(13);
\ENDFOR
\STATE Obtain the clustering results  $\mathbf{O}$ with the final $\mathbf{\widetilde{Z}}$ by K-means.
\RETURN $\mathbf{O}$
\end{algorithmic}}
\end{algorithm}

\subsection{Joint loss and Optimization}
The overall learning objective consists of two main parts, i.e., the reconstruction loss of AE and IGAE, and the clustering loss which is correlated with the target distribution:
\begin{equation} \label{eq:16}
 L= \underbrace{L_{AE}+L_{IGAE}}_{Reconstruction} +\underbrace{\lambda L_{KL}}_{Clustering}.
\end{equation}
In Eq.\eqref{eq:16}, $L_{AE}$ is the mean square error (MSE) reconstruction loss of AE. Different from SDCN, the proposed DFCN reconstructs the inputs of both sub-networks with the consensus latent representation. $\lambda$ is a pre-defined hyper-parameter which balances the importance of reconstruction and clustering. The detailed learning procedure of the proposed DFCN is shown in Algorithm 1.

\section{Experiments}
\subsection{Benchmark Datasets}
We evaluate the proposed DFCN on six popular public datasets, including three graph datasets (ACM$\footnote{ http://dl.acm.org/}$, DBLP$\footnote{https://dblp.uni-trier.de}$, and CITE$\footnote{ http://citeseerx.ist.psu.edu/index}$) and three non-graph datasets (USPS \cite{1990Handwritten}, HHAR \cite{2004RCV1}, and REUT \cite{2015Smart}). Table \ref{II} summarizes the brief information of these datasets. For the dataset (like USPS, HHAR, and REUT) whose affinity matrix is absent, we follow \cite{Bo2020Structural} and construct the matrix with heat kernel method. 

\subsection{Experiment Setup}
%Following \cite{Bo2020Structural}, Heat Kernel \cite{2009A} method is adopted to construct the KNN graph for non-graph datasets.

%In summary, we first individually pre-train both AE and IGAE to generate the DNN-based embedding $\mathbf{Z}_{AE}^{(m)}$ and GCN-based embedding $\mathbf{Z}_{IGAE}^{(n)}$ by optimizing \textit{L$_{AE}$} and \textit{L$_{IGAE}$}. Then $\mathbf{Z}_{AE}^{(m)}$ and $\mathbf{Z}_{IGAE}^{(n)}$ are directly fused to obtain the hybrid latent embedding in an united framework. Meanwhile, we perform a cross-modality dynamic fusion mechanism together with AE and IGAE to initialize the network parameters and clustering centroids \textit{u}. Thereafter, we calculate the shared-specific soft assignment distribution and target distribution via a triple self-supervised strategy according to Eq.\eqref{eq:13} and Eq.\eqref{eq:14}. Further, the model will achieve a stable convergence by minimizing the total loss Eq.\eqref{eq:16}, and we finally perform K-means method to achieve clustering based on the robust deep hybrid embedding $\mathbf{\widehat{Z}}$.

\subsubsection{Training Procedure}
Our method is implemented with PyTorch platform and a NVIDIA 2080TI GPU. The training of the proposed DFCN includes three steps. First, we pre-train the AE and IGAE independently for 30 iterations by minimizing the reconstruction loss functions. Then, both sub-networks are integrated into a united framework for another 100 iterations. Finally, with the learned centers of different clusters and under the guidance of the triplet self-supervised strategy, we train the whole network  for at least 200 iterations until convergence. The cluster ID is acquired by performing K-means algorithm over the consensus clustering embedding $\mathbf{\widetilde{Z}}$. Following all the compared methods, to alleviate the adverse influence of randomness, we repeat each experiment for 10 times and report the average values and the corresponding standard deviations.

\begin{table}[!t]
\centering

\footnotesize
%\label{tab:my-table}
\begin{tabular}{c|c|c|c|c}\hline
\hline
Dataset & Type   & Samples & Classes & Dimension \\\hline
USPS    & Image  & 9298    & 10      & 256       \\
HHAR    & Record & 10299   & 6       & 561       \\
REUT    & Text   & 10000   & 4       & 2000      \\
ACM     & Graph  & 3025    & 3       & 1870      \\
DBLP    & Graph  & 4058    & 4       & 334       \\
CITE    & Graph  & 3327    & 6       & 3703\\\hline\hline
\end{tabular}
\caption{Dataset summary}
\label{II}
\end{table}

\begin{table*}[!t]
\centering
\label{tab:my-table}
% \scriptsize
\tiny
\begin{tabular}{c|c|cccccccccc|c}
\hline\hline
Data               & Metric & K-means & AE    & DEC   & IDEC  & GAE   & VGAE  & ARGA & DAEGC & SDCN$_{Q}$& SDCN  & DFCN          \\\hline
\multirow{4}{*}{USPS} & ACC    & 66.8$\pm{0.0}$   & 71.0$\pm{0.0}$  & 73.3$\pm{0.2}$  & 76.2$\pm{0.1}$  & 63.1$\pm{0.3}$  & 56.2$\pm{0.7}$  &66.8$\pm{0.7}$& 73.6$\pm{0.4}$&77.1$\pm{0.2}$  & {\color{blue}78.1$\pm{0.2}$ } & {\color{red}79.5$\pm{0.2}$} \\
                      & NMI    & 62.6$\pm{0.0}$   & 67.5$\pm{0.0}$ & 70.6$\pm{0.3}$ & 75.6$\pm{0.1}$ & 60.7$\pm{0.6}$ & 51.1$\pm{0.4}$  &61.6$\pm{0.3}$ & 71.1$\pm{0.2}$     &77.7$\pm{0.2}$      & {\color{blue}79.5$\pm{0.3}$} & {\color{red}82.8$\pm{0.3}$} \\
                      & ARI    & 54.6$\pm{0.0}$   & 58.8$\pm{0.1}$ & 63.7$\pm{0.3}$ & 67.9$\pm{0.1}$ & 50.3$\pm{0.6}$ & 41.0$\pm{0.6}$ &51.1$\pm{0.6}$& 63.3$\pm{0.3}$     &70.2$\pm{0.2}$      & {\color{blue}71.8$\pm{0.2}$} & {\color{red}75.3$\pm{0.2}$} \\
                      & F1     & 64.8$\pm{0.0}$   & 69.7$\pm{0.0}$ & 71.8$\pm{0.2}$ & 74.6$\pm{0.1}$ & 61.8$\pm{0.4}$ & 53.6$\pm{1.1}$ &66.1$\pm{1.2}$ & 72.5$\pm{0.5}$  &75.9$\pm{0.2}$ & {\color{blue}77.0$\pm{0.2}$} & {\color{red}78.3$\pm{0.2}$}\\\hline
\multirow{4}{*}{HHAR} & ACC    & 60.0$\pm{0.0}$   & 68.7$\pm{0.3}$ & 69.4$\pm{0.3}$ & 71.1$\pm{0.4}$ & 62.3$\pm{1.0}$ & 71.3$\pm{0.4}$ &63.3$\pm{0.8}$ & 76.5$\pm{2.2}$      &83.5$\pm{0.2}$      & {\color{blue}84.3$\pm{0.2}$} & {\color{red}87.1$\pm{0.1}$} \\
                      & NMI    & 58.9$\pm{0.0}$   & 71.4$\pm{1.0}$ & 72.9$\pm{0.4}$ & 74.2$\pm{0.4}$ & 55.1$\pm{1.4}$ & 63.0$\pm{0.4}$ &57.1$\pm{1.4}$ & 69.1$\pm{2.3}$      &78.8$\pm{0.3}$      & {\color{blue}79.9$\pm{0.1}$} & {\color{red}82.2$\pm{0.1}$} \\
                      & ARI    & 46.1$\pm{0.0}$   & 60.4$\pm{0.9}$ & 61.3$\pm{0.5}$ & 62.8$\pm{0.5}$ & 42.6$\pm{1.6}$ & 51.5$\pm{0.7}$ &44.7$\pm{1.0}$ & 60.4$\pm{2.2}$      &71.8$\pm{0.2}$      & {\color{blue}72.8$\pm{0.1}$} & {\color{red}76.4$\pm{0.1}$} \\
                      & F1     & 58.3$\pm{0.0}$   & 66.4$\pm{0.3}$ & 67.3$\pm{0.3}$ & 68.6$\pm{0.3}$ & 62.6$\pm{1.0}$ & 71.6$\pm{0.3}$ &61.1$\pm{0.9}$ & 76.9$\pm{2.2}$      &81.5$\pm{0.1}$      & {\color{blue}82.6$\pm{0.1}$} & {\color{red}87.3$\pm{0.1}$} \\\hline
\multirow{4}{*}{REUT} & ACC    & 54.0$\pm{0.0}$   & 74.9$\pm{0.2}$ & 73.6$\pm{0.1}$ & 75.4$\pm{0.1}$ & 54.4$\pm{0.3}$ & 60.9$\pm{0.2}$ &56.2$\pm{0.2}$ & 65.6$\pm{0.1}$      &{\color{red}79.3$\pm{0.1}$}      & 77.2$\pm{0.2}$ & {\color{blue}77.7$\pm{0.2}$}\\
                      & NMI    & 41.5$\pm{0.5}$   & 49.7$\pm{0.3}$ & 47.5$\pm{0.3}$ & 50.3$\pm{0.2}$ & 25.9$\pm{0.4}$ & 25.5$\pm{0.2}$ &28.7$\pm{0.3}$  & 30.6$\pm{0.3}$     &{\color{blue}56.9$\pm{0.3}$}      & 50.8$\pm{0.2}$ & {\color{red}59.9$\pm{0.4}$} \\
                      & ARI    & 28.0$\pm{0.4}$   & 49.6$\pm{0.4}$ & 48.4$\pm{0.1}$ & 51.3$\pm{0.2}$ & 19.6$\pm{0.2}$ & 26.2$\pm{0.4}$  &24.5$\pm{0.4}$ & 31.1$\pm{0.2}$   &{\color{blue}59.6$\pm{0.3}$}      & 55.4$\pm{0.4}$ & {\color{red}59.8$\pm{0.4}$} \\
                      & F1     & 41.3$\pm{2.4}$   & 61.0$\pm{0.2}$ & 64.3$\pm{0.2}$ & 63.2$\pm{0.1}$ & 43.5$\pm{0.4}$ & 57.1$\pm{0.2}$ &51.1$\pm{0.2}$ & 61.8$\pm{0.1}$      &{\color{blue}66.2$\pm{0.2}$}      & 65.5$\pm{0.1}$ & {\color{red}69.6$\pm{0.1}$} \\\hline
\multirow{4}{*}{ACM}  & ACC    & 67.3$\pm{0.7}$   & 81.8$\pm{0.1}$ & 84.3$\pm{0.8}$ & 85.1$\pm{0.5}$ & 84.5$\pm{1.4}$ & 84.1$\pm{0.2}$ &86.1$\pm{1.2}$ & 86.9$\pm{2.8}$      &87.0$\pm{0.1}$      & {\color{blue}90.5$\pm{0.2}$} & {\color{red}90.9$\pm{0.2}$} \\
                      & NMI    & 32.4$\pm{0.5}$   & 49.3$\pm{0.2}$ & 54.5$\pm{1.5}$ & 56.6$\pm{1.2}$ & 55.4$\pm{1.9}$ & 53.2$\pm{0.5}$ &55.7$\pm{1.4}$ & 56.2$\pm{4.2}$     &58.9$\pm{0.2}$      & {\color{blue}68.3$\pm{0.3}$} & {\color{red}69.4$\pm{0.4}$} \\
                      & ARI    & 30.6$\pm{0.7}$   & 54.6$\pm{0.2}$ & 60.6$\pm{1.9}$ & 62.2$\pm{1.5}$ & 59.5$\pm{3.1}$ & 57.7$\pm{0.7}$ &62.9$\pm{2.1}$& 59.4$\pm{3.9}$      &65.3$\pm{0.2}$      & {\color{blue}73.9$\pm{0.4}$} & {\color{red}74.9$\pm{0.4}$} \\
                      & F1     & 67.6$\pm{0.7}$   & 82.0$\pm{0.1}$ & 84.5$\pm{0.7}$ & 85.1$\pm{0.5}$ & 84.7$\pm{1.3}$ & 84.2$\pm{0.2}$ &86.1$\pm{1.2}$ & 87.1$\pm{2.8}$     &86.8$\pm{0.1}$      & {\color{blue}90.4$\pm{0.2}$} & {\color{red}90.8$\pm{0.2}$} \\\hline
\multirow{4}{*}{DBLP} & ACC    & 38.7$\pm{0.7}$   & 51.4$\pm{0.4}$ & 58.2$\pm{0.6}$ & 60.3$\pm{0.6}$ & 61.2$\pm{1.2}$ & 58.6$\pm{0.1}$ &61.6$\pm{1.0}$ & 62.1$\pm{0.5}$      &65.7$\pm{1.3}$      & {\color{blue}68.1$\pm{1.8}$} & {\color{red}76.0$\pm{0.8}$} \\
                      & NMI    & 11.5$\pm{0.4}$   & 25.4$\pm{0.2}$ & 29.5$\pm{0.3}$  & 31.2$\pm{0.5}$ & 30.8$\pm{0.9}$ & 26.9$\pm{0.1}$ &26.8$\pm{1.0}$  & 32.5$\pm{0.5}$     &35.1$\pm{1.1}$      & {\color{blue}39.5$\pm{1.3}$} & {\color{red}43.7$\pm{1.0}$} \\
                      & ARI    & 7.0$\pm{0.4}$    & 12.2$\pm{0.4}$ & 23.9$\pm{0.4}$ & 25.4$\pm{0.6}$ & 22.0$\pm{1.4}$ & 17.9$\pm{0.1}$ &22.7$\pm{0.3}$& 21.0$\pm{0.5}$       &34.0$\pm{1.8}$      & {\color{blue}39.2$\pm{2.0}$} & {\color{red}47.0$\pm{1.5}$} \\
                      & F1     & 31.9$\pm{0.3}$   & 52.5$\pm{0.4}$ & 59.4$\pm{0.5}$ & 61.3$\pm{0.6}$ & 61.4$\pm{2.2}$ & 58.7$\pm{0.1}$  &61.8$\pm{0.9}$& 61.8$\pm{0.7}$     &65.8$\pm{1.2}$      & {\color{blue}67.7$\pm{1.5}$} & {\color{red}75.7$\pm{0.8}$} \\\hline
\multirow{4}{*}{CITE} & ACC    & 39.3$\pm{3.2}$   & 57.1$\pm{0.1}$ & 55.9$\pm{0.2}$ & 60.5$\pm{1.4}$ & 61.4$\pm{0.8}$ & 61.0$\pm{0.4}$  &56.9$\pm{0.7}$& 64.5$\pm{1.4}$    &61.7$\pm{1.1}$      & {\color{blue}66.0$\pm{0.3}$} & {\color{red}69.5$\pm{0.2}$} \\
                      & NMI    & 16.9$\pm{3.2}$   & 27.6$\pm{0.1}$ & 28.3$\pm{0.3}$ & 27.2$\pm{2.4}$ & 34.6$\pm{0.7}$ & 32.7$\pm{0.3}$ &34.5$\pm{0.8}$& 36.4$\pm{0.9}$       &34.4$\pm{1.2}$      & {\color{blue}38.7$\pm{0.3}$} & {\color{red}43.9$\pm{0.2}$} \\
                      & ARI    & 13.4$\pm{3.0}$   & 29.3$\pm{0.1}$ & 28.1$\pm{0.4}$ & 25.7$\pm{2.7}$ & 33.6$\pm{1.2}$ & 33.1$\pm{0.5}$ &33.4$\pm{1.5}$  & 37.8$\pm{1.2}$    &35.5$\pm{1.5}$      & {\color{blue}40.2$\pm{0.4}$} & {\color{red}45.5$\pm{0.3}$} \\
                      & F1     & 36.1$\pm{3.5}$   & 53.8$\pm{0.1}$ & 52.6$\pm{0.2}$ & 61.6$\pm{1.4}$ & 57.4$\pm{0.8}$ & 57.7$\pm{0.5}$&54.8$\pm{0.8}$ & 62.2$\pm{1.3}$      &57.8$\pm{1.0}$      & {\color{blue}63.6$\pm{0.2}$} & {\color{red}64.3$\pm{0.2}$} \\\hline\hline
\end{tabular}
\caption{Clustering performance on six datasets (mean$\pm{}$std).  
The {\color{red}red} and {\color{blue}blue} values indicate the best and the runner-up results, respectively.}
\label{III}

%\end{sidewaystable}
\end{table*}

\subsubsection{Parameters Setting}
For ARGA \cite{Pan2019Learning}, we set the parameters of the method by following the setting of the original paper. For other compared methods, we report the results listed in the paper SDCN \cite{Bo2020Structural} directly. For our method, we adopt the original code and data of SDCN for data pre-processing and testing.
%and further adopt the Principal Components Analysis (PCA) to extract more compact features as inputs. 
All ablation studies are trained with the Adam optimizer. The optimization stops when the validation loss comes to a plateau. The learning rate is set to 1e-3 for USPS, HHAR, 1e-4 for REUT, DBLP, and CITE, and 5e-5 for ACM. The training batch size is set to 256 and we adopt an early stop strategy to avoid over-fitting. According to the results of parameter sensitivity testing, we fix two balanced hyper-parameters $\gamma$ and $\lambda$ to 0.1 and 10, respectively. Moreover, we set the nearest neighbors number of each node as 5 for all non-graph datasets.

%The dimension of AE and IGAE are both set as \textit{d}-128-256-512-20 for the encoder and \textit{d}-128-256-20 for the decoder in all the experiment.
\subsubsection{Evaluation Metric}
The clustering performance of all methods is evaluated by four metrics: Accuracy (ACC), Normalized Mutual Information (NMI), Average Rand Index (ARI), and macro F1-score (F1) \cite{zhou2020subspace,zhou2019multiple,2020Absent,2020Multiple,2019Late}. The best map between cluster ID and class ID is found by using the Kuhn-Munkres algorithm \cite{1986Matching}.

\subsection{Comparison with the State-of-the-art Methods}
In this part, we compare our proposed method with ten state-of-the-art clustering methods to illustrate its effectiveness. Among them, K-means \cite{1979A} is the representative one of classic shallow clustering methods. AE \cite{ae}, DEC \cite{2015Unsupervised}, and IDEC \cite{2017Improved}  represent the autoencoder-based clustering methods which learn the representations for clustering through training an autoencoder. GAE/VGAE \cite{2016Variational}, ARGA \cite{Pan2019Learning}, and DAEGC \cite{DAEGC} are typical methods of graph convolutional network-based methods. In these methods, the clustering representation is embedded with structure information by GCN. SDCN$_{Q}$ and SDCN \cite{Bo2020Structural} are representatives of hybrid methods which take advantage of both AE and GCN module for clustering.

\begin{figure}[!t]
\centering
\includegraphics[width=3.1in]{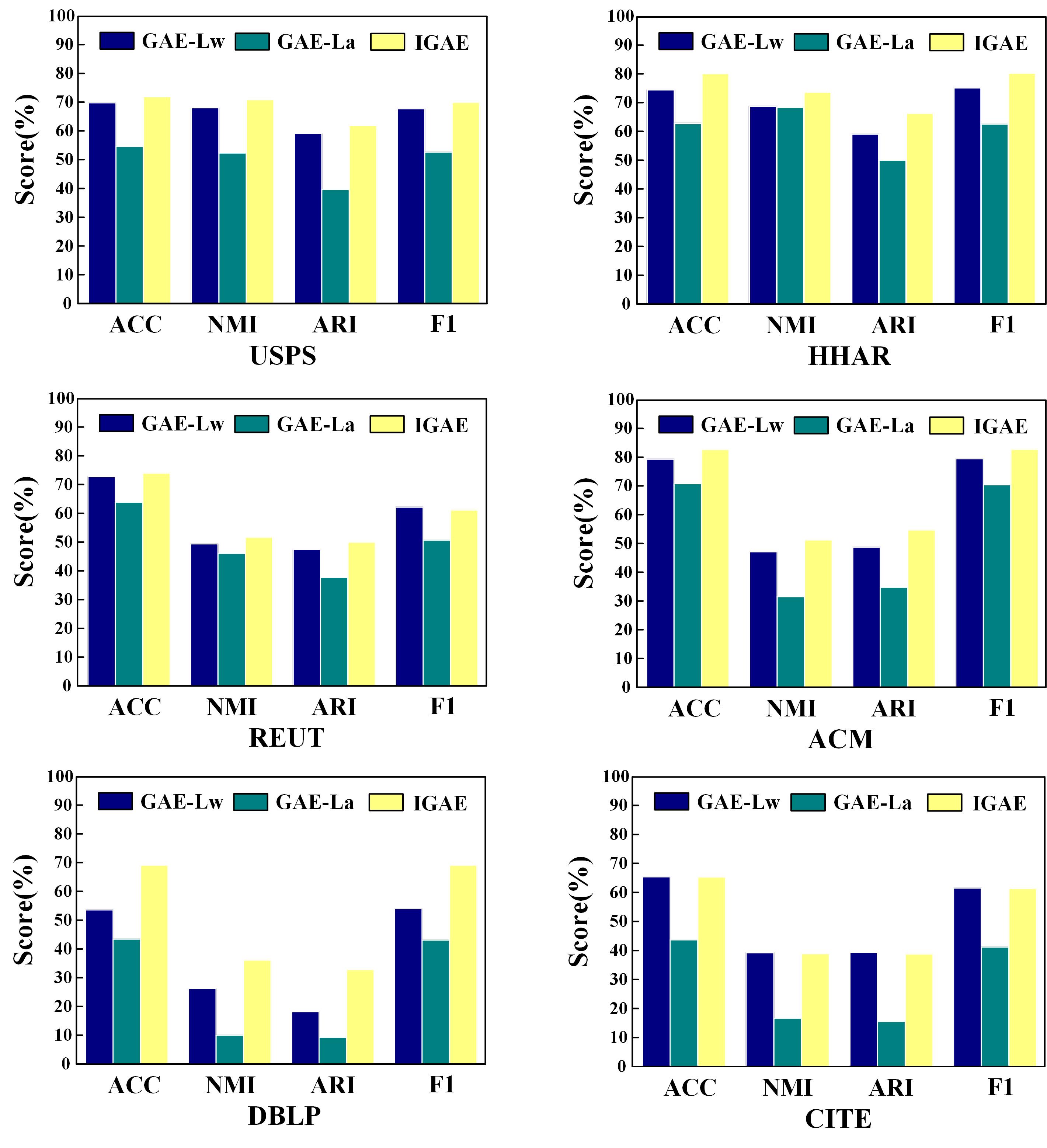}
% where an .eps filename suffix will be assumed under latex,
% and a .pdf suffix will be assumed for pdflatex; or what has been declared
% via \DeclareGraphicsExtensions.
\caption{Clustering results of the graph autoencoder with different reconstruction strategy. GAE-L$_{w}$, GAE-L$_{a}$, and IGAE correspond to the reconstruction of weighted attribute matrix, adjacency matrix, and both.}
\label{2}
\end{figure}

\begin{figure*}[!t]
\centering
\includegraphics[width=7.02in]{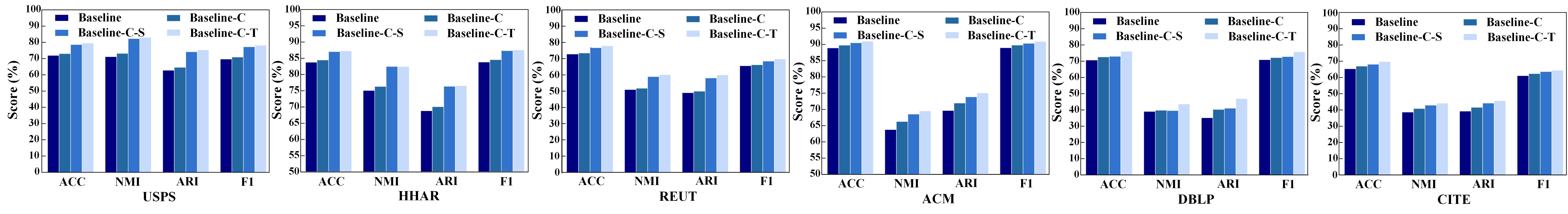}
% where an .eps filename suffix will be assumed under latex,
% and a .pdf suffix will be assumed for pdflatex; or what has been declared
% via \DeclareGraphicsExtensions.
\caption{Ablation comparisons of cross-modality dynamic fusion mechanism and triplet self-supervised strategy in SAIF. The baseline refers to a naive united framework consisting of AE and IGAE. -C, -S, and -T indicate that the baseline utilizes the cross-modality dynamic fusion mechanism, single or triplet self-supervised strategy, respectively.}
\label{3}
\end{figure*}

The clustering performance of our method and 10 baseline methods on six benchmark datasets are summarized in Table \ref{III}. Based on the results, we have the following observations:

1) DFCN shows superior performance against the compared methods in most circumstances. Specifically, K-means performs clustering on raw data. AE, DEC, and IDEC merely exploit node attribute representations for clustering. These methods seldom take structure information into account, leading to sub-optimal performance. In contrast, DFCN successfully leverages available data by selectively integrating the information of graph structure and node attributes, which complements each other for consensus representation learning and greatly improves clustering performance.

2) It is obvious that GCN-based methods such as GAE, VGAE, ARGA, and DAEGC are not comparable to ours, because these methods under-utilize abundant information from data itself and might be limited to the over-smoothing phenomenon. Differently, DFCN incorporates attribute-based representations learned by AE into the whole clustering framework, and mutually explores graph structure and node attributes with a fusion module for consensus representation learning. As a result, the proposed DFCN improves the clustering performance of the existing GCN-based methods with a preferable gap.

%contributes to significant improvements on clustering performance compared with the above methods.

3) DFCN achieves better clustering results than the strongest baseline methods SDCN$_{Q}$ and SDCN in the majority of cases, especially on HHAR, DBLP, and CITE datasets. On DBLP dataset for instance, our method achieves a  7.9\%, 4.2\%, 7.8\%, and 8.0\% increment with respect to ACC, NMI, ARI and F1 against SDCN. This is because DFCN not only achieves a dynamic interaction between graph structure and node attributes to reveal the intrinsic clustering structure, but also adopts a triplet self-supervised strategy to provide precise network training guidance.

\begin{table}[!t]
\centering
\scriptsize
% \tiny
\begin{tabular}{c|c|cccc}
  \hline
  \hline
  Dataset&Model&ACC&NMI&ARI&F1\\
  \hline

\multirow{3}{*}{USPS} & +AE   & 78.3$\pm{0.3}$ & 81.3$\pm{0.1}$ & 73.6$\pm{0.3}$ & 76.8$\pm{0.3}$\\
                      & +IGAE & 76.9$\pm{0.4}$ & 77.1$\pm{0.4}$ & 68.8$\pm{0.6}$ & 74.8$\pm{0.5}$\\
                      & DFCN  & \textbf{79.5$\pm{\textbf{0.2}}$} & \textbf{82.8$\pm{\textbf{0.3}}$} & \textbf{75.3$\pm{\textbf{0.2}}$} & \textbf{78.3$\pm{\textbf{0.2}}$} \\ \hline
\multirow{3}{*}{HHAR} & +AE   & 75.2$\pm{1.4}$& \textbf{82.8}$\pm{\textbf{1.0}}$& 71.7$\pm{1.2}$& 72.6$\pm{0.9}$\\
                      & +IGAE & 82.8$\pm{0.1}$& 79.6$\pm{0.1}$& 72.3$\pm{0.1}$& 83.4$\pm{0.1}$\\
                      & DFCN  & \textbf{87.1$\pm{\textbf{0.1}}$} & 82.2$\pm{0.1}$ & \textbf{76.4$\pm{\textbf{0.1}}$} & \textbf{87.3$\pm{\textbf{0.1}}$} \\ \hline
\multirow{3}{*}{REUT} & +AE   & 69.3$\pm{0.8}$& 48.5$\pm{1.6}$& 44.6$\pm{1.1}$& 58.3$\pm{0.6}$\\
                      & +IGAE & 71.4$\pm{1.7}$& 52.5$\pm{1.0}$& 49.1$\pm{2.2}$& 61.5$\pm{2.9}$\\
                      & DFCN  & \textbf{77.7$\pm{\textbf{0.2}}$} & \textbf{59.9$\pm{\textbf{0.4}}$} & \textbf{59.8$\pm{\textbf{0.4}}$} & \textbf{69.6$\pm{\textbf{0.1}}$} \\ \hline
\multirow{3}{*}{ACM}  & +AE   & 90.2$\pm{0.3}$& 67.5$\pm{0.8}$& 73.2$\pm{0.8}$& 90.2$\pm{0.3}$\\
                      & +IGAE & 89.6$\pm{0.2}$& 65.6$\pm{0.4}$& 71.8$\pm{0.4}$& 89.6$\pm{0.2}$\\
                      & DFCN  & \textbf{90.9$\pm{\textbf{0.2}}$} & \textbf{69.4$\pm{\textbf{0.4}}$} & \textbf{74.9$\pm{\textbf{0.4}}$} & \textbf{90.8$\pm{\textbf{0.2}}$} \\ \hline
\multirow{3}{*}{DBLP} & +AE   & 64.2$\pm{2.9}$& 30.2$\pm{3.2}$& 29.4$\pm{3.4}$& 64.6$\pm{2.8}$\\
                      & +IGAE & 67.5$\pm{1.0}$& 34.2$\pm{1.1}$& 31.5$\pm{1.1}$& 67.6$\pm{1.0}$\\
                      & DFCN  & \textbf{76.0$\pm{\textbf{0.8}}$} & \textbf{43.7$\pm{\textbf{1.0}}$} & \textbf{47.0$\pm{\textbf{1.5}}$} & \textbf{75.7$\pm{\textbf{0.8}}$} \\ \hline
\multirow{3}{*}{CITE} & +AE   & 69.3$\pm{0.3}$& 42.9$\pm{0.4}$& 44.7$\pm{0.4}$& \textbf{64.4$\pm{\textbf{0.3}}$}\\
                      & +IGAE & 67.9$\pm{0.9}$& 41.8$\pm{1.0}$& 43.0$\pm{1.4}$& 63.7$\pm{0.7}$\\
                      & DFCN  & \textbf{69.5$\pm{\textbf{0.2}}$} & \textbf{43.9$\pm{\textbf{0.2}}$} & \textbf{45.5$\pm{\textbf{0.3}}$} & 64.3$\pm{0.2}$ \\ \hline\hline
\end{tabular}
\caption{Ablation comparisons of the target distribution generation with signle- or both-source information.}
\label{V}
\end{table}

\subsection{Ablation Studies}
\subsubsection{Effectiveness of IGAE}
We further conduct ablation studies to verify the effectiveness of IGAE and report the results in Fig. \ref{2}. GAE-L$_{w}$ or GAE-L$_{a}$ denotes the method optimized by the reconstruction loss function of weighted attribute matrix or adjacency matrix only. We can find out that GAE-L$_{w}$ consistently performs better than GAE-L$_{a}$ on six datasets. Besides, IGAE clearly improves the clustering performance over the method which constructs the adjacency matrix only. Both observations illustrate that our proposed reconstruction measure is able to exploit more comprehensive information for improving the generalization capability of the deep clustering network. 
By this means, the latent embedding inherits more properties from the attribute space of the original graph, preserving representative features that generate better clustering decisions.

\begin{figure}[!t]
\centering
\includegraphics[width=3.3in]{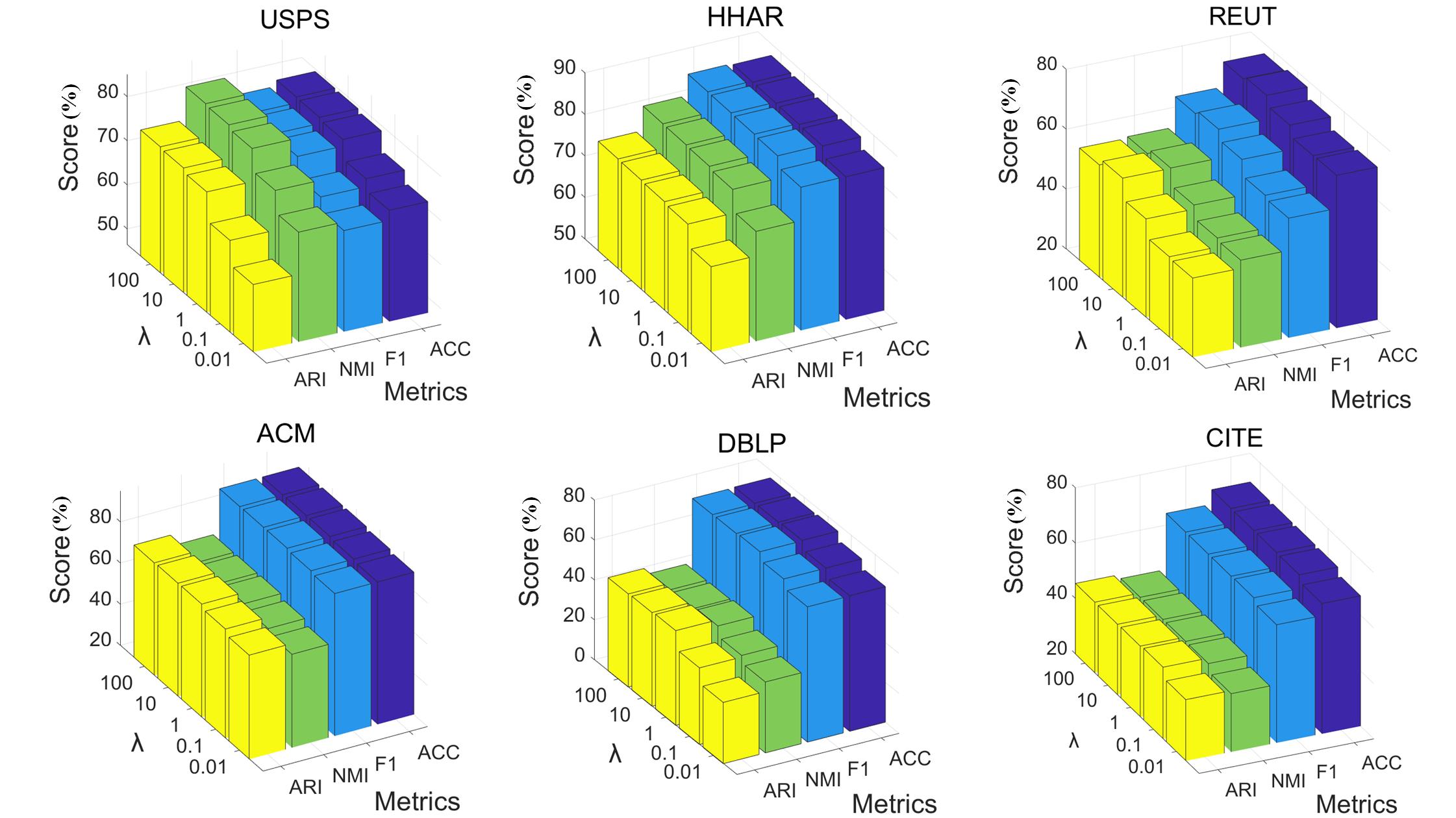}
% where an .eps filename suffix will be assumed under latex,
% and a .pdf suffix will be assumed for pdflatex; or what has been declared
% via \DeclareGraphicsExtensions.
\caption{The sensitivity of DFCN with the variation of $\lambda$ on six datasets.} %Colors indicate different clusters.}
\label{6}
\end{figure}

%\begin{figure*}[!t]
%\centering
%\includegraphics[width=6.5in]{f13.jpg}
% where an .eps filename suffix will be assumed under latex,
% and a .pdf suffix will be assumed for pdflatex; or what has been declared
% via \DeclareGraphicsExtensions.
%\caption{XXXX} %Colors indicate different clusters.}
%\label{5}
%\end{figure*}

\begin{figure*}[!t]
\centering
\includegraphics[width=6.5in]{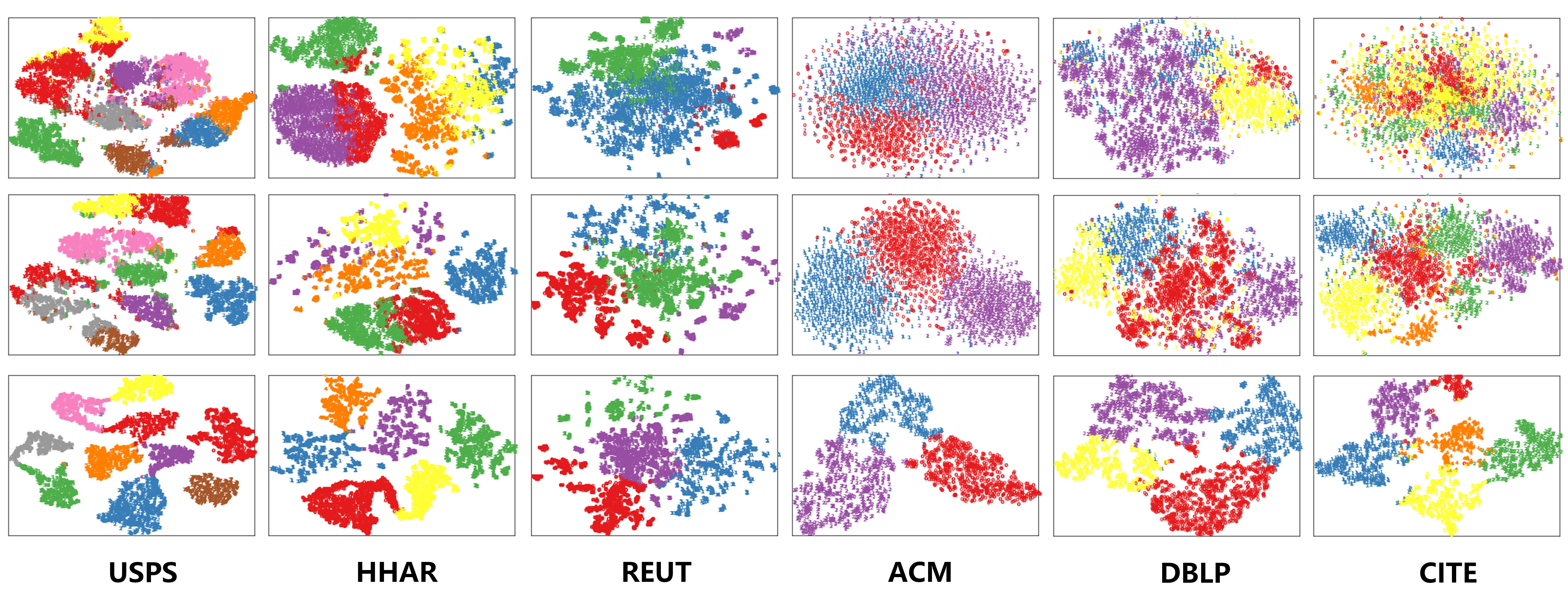}
% where an .eps filename suffix will be assumed under latex,
% and a .pdf suffix will be assumed for pdflatex; or what has been declared
% via \DeclareGraphicsExtensions.
\caption{2D visualization on six datasets. The first, second, and last row correspond to the distribution of raw data, baseline and DFCN (baseline + SAIF), respectively.} %Colors indicate different clusters.}
\label{5}
\end{figure*}

\subsubsection{Analysis of the SAIF Module}
In this part, we conduct several experiments to verify the effectiveness of the SAIF module. As summarized in Fig. \ref{3}, we observe that 1) compared with the baseline, Baseline-C method has about 0.5\% to 5.0\% performance improvements, indicating that exploring graph structure and node attributes in both the perspective of the local and global level is helpful to learn consensus latent representations for better clustering; 2) the performance of Baseline-C-T method is consistently better than that of Baseline-C-S method on all datasets. The reason is that our triplet self-supervised strategy successfully generates more reliable guidance for the training of AE, IGAE, and the fusion part, making them benefit from each other. According to these observations, the superiority of the SAIF module has clearly been demonstrated over the baseline.

\subsubsection{Influence of Exploiting Both-source Information}
We compare our method with two variants to validate the effectiveness of complementary two-modality (structure and attribute) information learning for target distribution generation. As reported in Table \ref{V}, +AE or +IGAE refers to the DFCN with only AE or IGAE part, respectively. On one hand, as +AE and +IGRE achieve better performance on different datasets, it indicates that information from either AE or IGAE cannot consistently outperform that of their counterparts, combining the both-source information can potentially improve the robustness of the hybrid method. On the other hand, DFCN encodes both DNN- and GCN-based representations and consistently outperforms the single-source methods. This shows that 1) both-source information is equally essential for the performance improvement of DFCN; 2) DFCN can facilitate the complementary two-modality information to make the target distribution more reliable and robust for better clustering.

\subsection{Analysis of Hyper-parameter $\lambda$}
As can be seen in Eq.\eqref{eq:16}, DFCN introduces a hyper-parameter $\lambda$ to make a trade-off between the reconstruction and clustering. We conduct experiments to show the effect of this parameter on all datasets. Fig. \ref{6} illustrates the performance variation of DFCN when $\lambda$ varies from 0.01 to 100. From these figures, we observe that 1) the hyper-parameter $\lambda$ is effective in improving the clustering performance; 2) the performance of the method is stable in a wide range of $\lambda$; 3) DFCN tends to perform well by setting $\lambda$ to 10 across all datasets.

\subsection{Visualization of Clustering Results}
To intuitively verify the effectiveness of DFCN, we visualize the distribution of the learned clustering embedding $\mathbf{\widetilde{Z}}$ in two-dimensional space by employing \textit{t}-SNE algorithm~\cite{2008Visualizing}. As illustrated in Fig. \ref{5}, %we can observe that our baseline can well cluster the samples according to their corresponding centroids, indicating that the deep hybrid embedding is meaningful. Additionally, 
DFCN can better reveal the intrinsic clustering structure among data.

% Appendix C contains additional experimental results, including the effect of over-smoothing problem by varying the depth of IGAE, the number of nearest neighbors \textit{K}-sensitivity analysis in the construction of KNN graph, and the convergence visualization of the proposed method.

\section{Conclusion}
In this paper, we propose a novel neural network-based clustering method termed Deep Fusion Clustering Network (DFCN). In our method, the core component SAIF module leverages both graph structure and node attributes via a dynamic cross-modality fusion mechanism and a triplet self-supervised strategy. In this way, more consensus and discriminative information from both sides is encoded to construct the robust target distribution, which effectively provides the precise network training guidance. Moreover, the proposed IGAE is able to assist in improving the generalization capability of the proposed method. Experiments on six benchmark datasets show that DFCN consistently outperforms state-of-the-art baseline methods. In the future, we plan to further improve our method to adapt it to multi-view graph clustering and incomplete multi-view graph clustering applications.

%The IGAE is designed to incorporate graph itself as regularized information into the reconstruction process, thus boosting the graph embedding effectively.

\section{Acknowledgments}
\bigskip
\noindent This work is supported by the National Key R $\&$ D Program of China (Grant 2018YFB1800202, 2020AAA0107100, 2020YFC2003400), the National Natural Science Foundation of China (Grant 61762033, 62006237, 62072465), the Hainan Province Key R $\&$ D Plan Project (Grant ZDYF2020040), the Hainan Provincial Natural Science Foundation of China (Grant 2019RC041, 2019RC098), and the Opening Project of Shanghai Trusted Industrial Control Platform (Grant TICPSH202003005-ZC).

% , and the Education and Teaching Reform Research Project of Hainan University under Grant hdjy1970.

\bibliography{myAAAIBib}
\bibliographystyle{aaai}

\end{document}